%% file: main.tex
\pdfoutput=1
\documentclass[11pt]{article}
\usepackage[preprint]{acl}
\usepackage{times}
\usepackage{latexsym}
\usepackage[T1]{fontenc}
\usepackage[utf8]{inputenc}
\usepackage{microtype}
\usepackage{inconsolata}
\usepackage{graphicx}
\usepackage{times}
\usepackage{latexsym}
\usepackage{amsmath}
\usepackage{amssymb}
\usepackage{enumitem}
\usepackage{graphicx}
\usepackage{subcaption}
\usepackage{pifont}
\usepackage{xcolor}
\usepackage{titletoc}
\usepackage{colortbl}
\usepackage{wrapfig}  
\usepackage{lipsum} 
\usepackage{multirow}
\usepackage{listings}
\usepackage{array}
\usepackage{booktabs}
\usepackage{caption}  
\usepackage{multicol}
\usepackage{xurl}
\usepackage{inconsolata}
\usepackage{ulem}
\usepackage{array}
\usepackage[ruled,linesnumbered,noend]{algorithm2e}
\usepackage{algpseudocode}
\usepackage{float}
\usepackage[most]{tcolorbox}
\usepackage{authblk}

\makeatletter
\newcommand{\nofootnote}[1]{%
    \begingroup
    \renewcommand\@makefntext[1]{\parindent 1em\noindent##1}
    \footnotetext{#1}% 
    \endgroup
}
\makeatother

\newcommand{\ours}{EduVisBench}
\newcommand{\ma}{EduVisAgent}
\hypersetup{
    colorlinks=true,
    linkcolor=red,
    citecolor=blue,
    filecolor=magenta,      
    urlcolor=blue,
linktocpage}

\title{From \ours\ to \ma: A Benchmark and Multi-Agent Framework for Reasoning-Driven Pedagogical Visualization}

\makeatletter
\newcommand{\equalfootnote}{%
  \gdef\@thefnmark{*}\@footnotetext{Equal contribution. Correspondence to: \{haonianj, huaxiu\}@cs.unc.edu}%
}
\makeatother

\author{
  \textbf{Haonian Ji$^{1}$\textsuperscript{*}, Shi Qiu$^{1}$\textsuperscript{*}, Siyang Xin$^{1}$\textsuperscript{*}, Siwei Han$^{1}$}\textsuperscript{*} \\
  \textbf{Zhaorun Chen$^{2}$, Dake Zhang$^{3}$, Hongyi Wang$^{3}$, Huaxiu Yao$^{1}$} \\
  $^{1}$UNC-Chapel Hill, $^{2}$University of Chicago, $^{3}$Rutgers University
}

\begin{document}
\maketitle
\equalfootnote

\input{sec/abs}
\input{sec/intro}
\input{sec/method}
\input{sec/multiagent}

\input{sec/experiment}
\input{sec/related-work}

\input{sec/conclusion}

\bibliography{custom}

\input{sec/appendix}

\end{document}

%% file: sec/abs.tex
\begin{abstract}

While foundation models (FMs), such as diffusion models and large vision-language models (LVLMs), have been widely applied in educational contexts, their ability to generate pedagogically effective visual explanations remains limited. Most existing approaches focus primarily on textual reasoning, overlooking the critical role of structured and interpretable visualizations in supporting conceptual understanding. To better assess the visual reasoning capabilities of FMs in educational settings, we introduce \ours, a multi-domain, multi-level benchmark. \ours\ features diverse STEM problem sets requiring visually grounded solutions, along with a fine-grained evaluation rubric informed by pedagogical theory. Our empirical analysis reveals that existing models frequently struggle with the inherent challenge of decomposing complex reasoning and translating it into visual representations aligned with human cognitive processes. To address these limitations, we propose \ma, a multi-agent collaborative framework that coordinates specialized agents for instructional planning, reasoning decomposition, metacognitive prompting, and visualization design. Experimental results show that \ma\ substantially outperforms all baselines, achieving a 40.2\% improvement and delivering more educationally aligned visualizations. \ours\ and \ma\ are available at \href{https://github.com/aiming-lab/EduVisBench}{github.com/aiming-lab/EduVisBench} and \href{https://github.com/aiming-lab/EduVisAgent}{github.com/aiming-lab/EduVisAgent}.

\end{abstract}

%% file: sec/intro.tex
\section{Introduction}
\label{intro}
\textit{``To truly teach is not to tell the answer, but to illuminate the path."}
\\

\vspace{-1em}

While foundation models (FMs), such as diffusion models and large vision-language models (LVLMs), have been extensively adopted in educational domains~\cite{chu2025llm, wang2024large}, including pedagogical agents providing automated classroom assistance and science learning agents offering textual explanations of problem-solving processes~\cite{wu2023mathchat}, their applications have predominantly focused on text-based interactions~\cite{wu2023mathchat, xu2024eduagent}.
However, in education, especially K-12 settings, creating compelling visualizations is crucial for cognitive comprehension and overall learning effectiveness ~\cite{presmeg2006research}. Despite its importance, there is currently limited understanding of how FMs can effectively generate visually grounded elements (e.g., \textit{diagrams}, \textit{interactive education tools}, \textit{illustrative graphics}) to support the pedagogical illustration of problem-solving processes.

\begin{figure}[t!]
    \centering
    \includegraphics[width=0.8\linewidth]{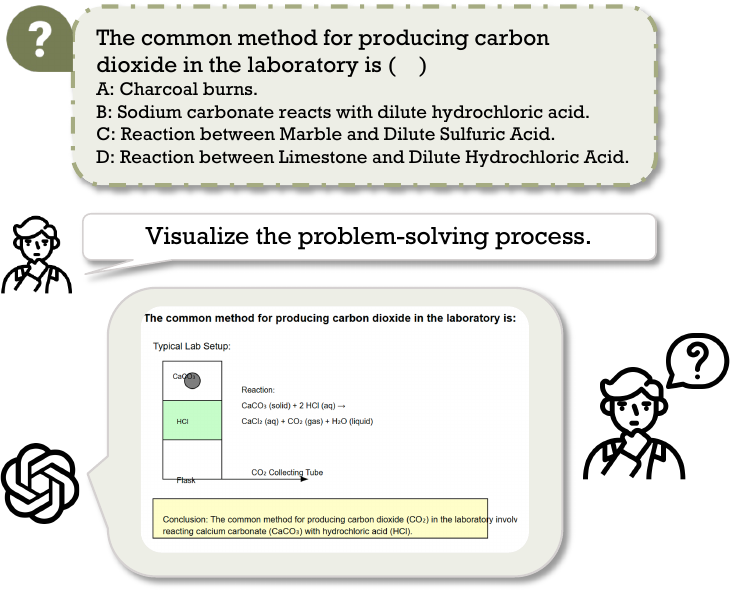}
    \caption{GPT-4o fails to illustrate its problem-solving with high-quality, logical, and explanatory visualization.}
    \label{fig:teaser}
    \vspace{-1.5em}
\end{figure}

Currently, generating visually grounded elements for pedagogical reasoning poses several challenges: (1) decomposing complex reasoning into representable steps that align closely with human cognitive processes is non-trivial~\cite{yang2024matplotagent, chen2024autoprm}; (2) precisely producing visual aids for each sub-step to optimally support learners is challenging~\cite{hong2025llms}; and (3) different educational domains require distinct visualization styles and formats, making consistent and adequate visual aid delivery difficult~\cite{pandey2025benchmarking}. Addressing these obstacles first requires a clear picture of how current FMs perform, so that future models can be purpose-built to close the gaps. Consequently, a comprehensive evaluation platform is critical for systematically assessing FMs on visual pedagogical reasoning.

To bridge this gap, we introduce \textbf{\ours}, a multi-domain, multi-level benchmark designed to evaluate the capacity of foundation models to generate pedagogically effective, step-by-step visual reasoning. \ours\ comprises structured problem sets across diverse domains, each requiring multimodal-centric reasoning and solutions that prioritize visualization principles such as \textit{interpretability}, \textit{cognitive alignment}, and \textit{instructional clarity} to achieve high evaluation score. To facilitate a detailed evaluation, we further develop a fine-grained rubric enabling multidimensional assessments of AI-generated visual outputs, focusing explicitly on pedagogical criteria such as \textit{contextual relevance}, \textit{visual clarity}, \textit{multimodal coherence}, \textit{reasoning support}, and \textit{interactive engagement}.

Utilizing this benchmark, we conduct extensive evaluations on a variety of FMs and agents. Our findings reveal that although current models achieve predominantly correct step-by-step textual analyses, they frequently fail to generate useful or faithful visualizations, as depicted in Figure~\ref{fig:teaser}. Specifically, our systematic analysis highlights recurring challenges including (1) semantic misalignments between textual explanations and visual components, (2) omissions of critical steps within rendered diagrams, and (3) structural inconsistencies in code-based visual outputs, collectively undermining accuracy, clarity, and interactivity.

To address these limitations, we introduce a multi-agent collaborative framework, \textbf{\ma}, designed to simulate the complete learning journey—from initial problem exposure to deep conceptual understanding. Specifically, a central planning agent orchestrates six specialized expert agents dedicated to \textit{visualization design}, \textit{cognitive scaffolding}, and \textit{metacognitive regulation}. A synthesis module then integrates these expert outputs into interactive, personalized learning webpages tailored specifically to human learners. Experimental results demonstrate that our proposed method \ma\ achieves an average improvement of 40.2\% than current SOTA method. This underscores the effectiveness of our approach—leveraging modular specialization and collaborative integration to produce robust and visually grounded learning solutions. 

% Furthermore, XXX.

\begin{figure}[t!]
    \centering
    \includegraphics[width=0.8\linewidth]{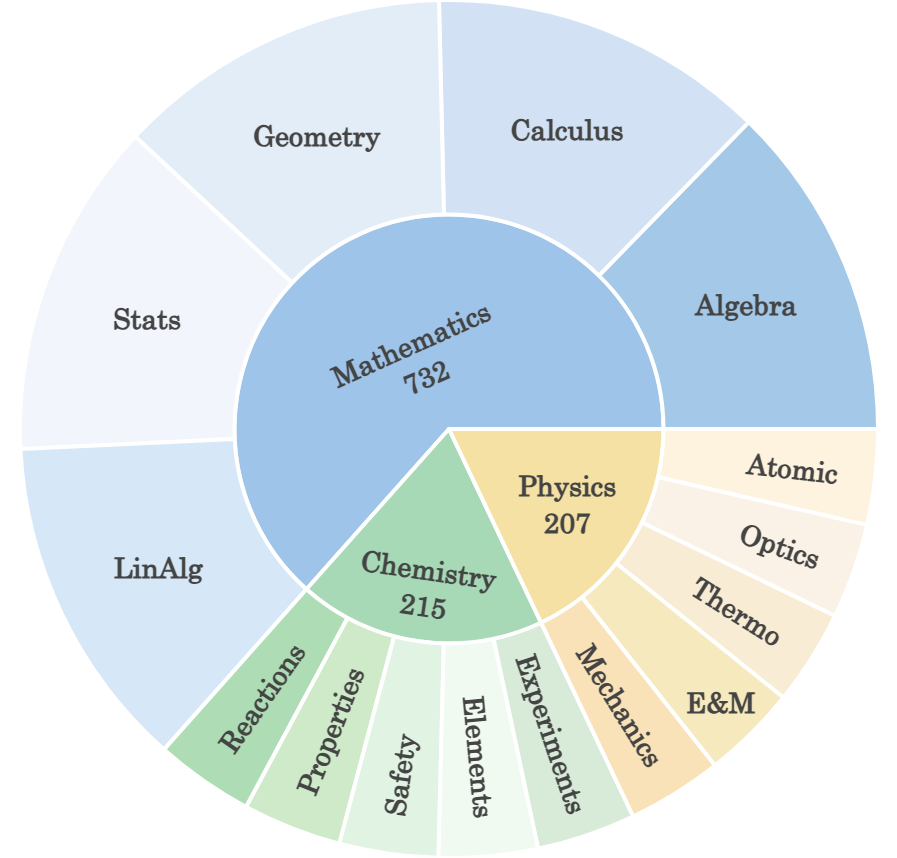}
    \caption{Dataset distribution of \ours. Each domain encompasses various sub-domains, collectively covering 15 comprehensive pedagogical scenarios.}
    \label{fig:distribution}
    \vspace{-1.5em}
\end{figure}

\begin{figure*}[htp]
    \centering
    \includegraphics[width=0.95\linewidth]{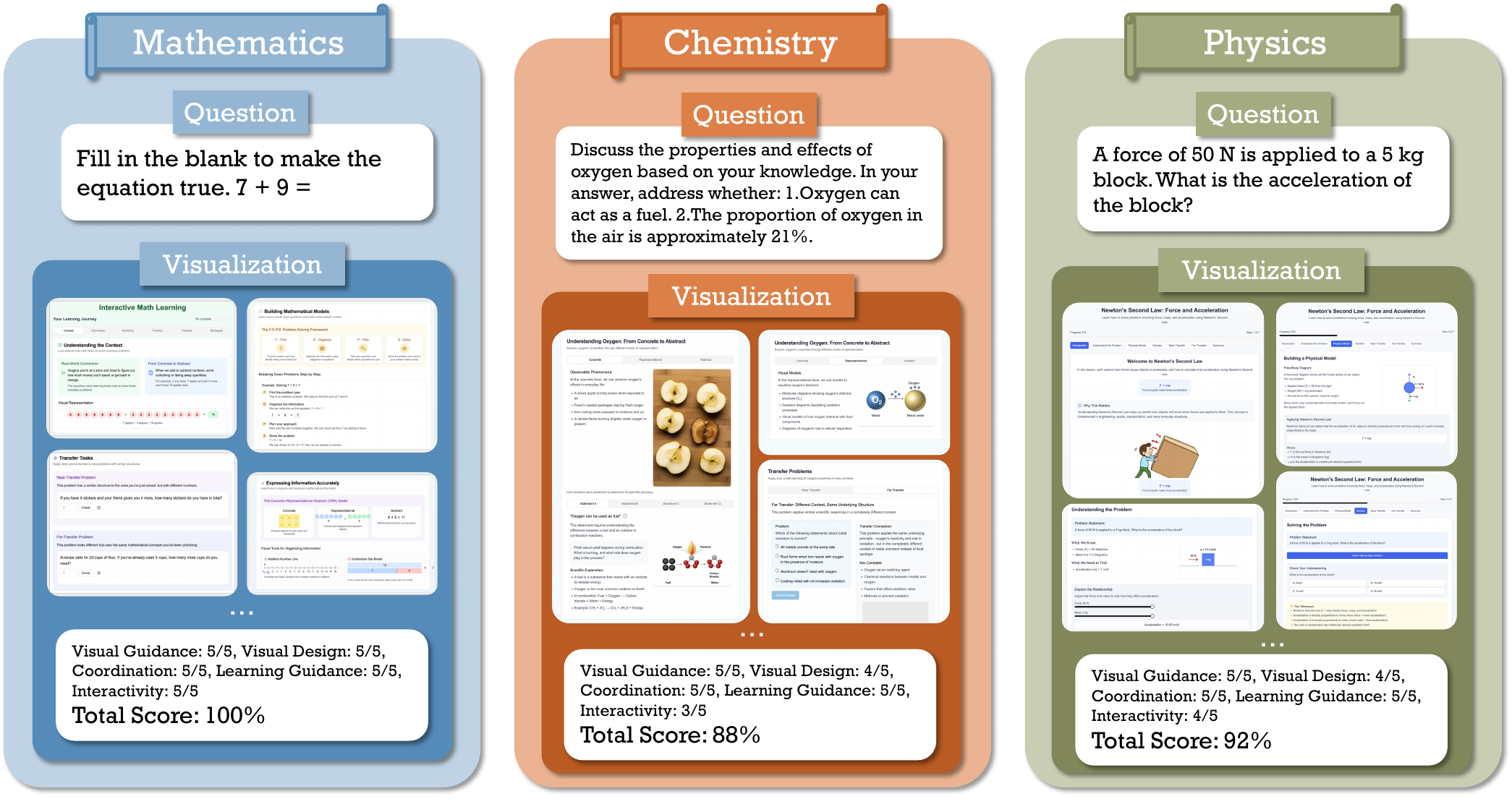}
    \caption{Representative examples from \ours, featuring questions from Maths, Chemistry, and Physics alongside their corresponding high-scoring visual explanations. These interactive visualizations, generated by our multi-agent system \ma, exemplify well-designed, pedagogically effective outputs for STEM problems.}
    \label{fig:main-figure}
    \vspace{-1em}
\end{figure*}

%% file: sec/method.tex
\section{\ours\ Benchmark}
\label{benchmark}

\vspace{-0.1in}

\subsection{Overview}
\vspace{-0.05in}

In this section, we introduce \ours, a novel and challenging benchmark designed to evaluate the capability of models to generate logical and explanatory visualizations for educational purposes. As shown in Figure~\ref{fig:distribution}, \ours\ comprises 1,154 carefully curated STEM questions across three academic subjects and 15 distinct domains, organized into three levels of difficulty. In addition to assessing accuracy in step-by-step problem solving, \ours\ places particular emphasis on a model’s ability to communicate the reasoning process clearly and visually—helping students understand problems through structured, interpretable visual outputs, as illustrated in Figure~\ref{fig:main-figure}.

Specifically, \ours\ adopts a multimodal setting in which models are provided with both textual and visual inputs and are tasked with producing diverse output formats, including interactive web pages and visual diagrams. Beyond evaluating the correctness of final answers, we introduce a fine-grained evaluation framework that assesses the quality of visualizations across five key dimensions: (1) the logical sequencing of visual elements, (2) the structural richness of the visuals, (3) semantic alignment with the underlying subject matter, (4) the clarity and guidance provided for problem-solving, and (5) the level of interactivity and engagement. In the following subsections, we describe our dataset curation process and the design of the evaluation rubric in detail.

\subsection{Dataset Curation}  
\label{sec:curation}  

\ours\ is built from several high-quality public educational resources that we carefully curated, translated, and adapted to support multimodal visualization learning tasks. Specifically, the chemistry questions are sourced from the \textbf{C-MHChem-Benchmark}~\cite{zhang2024chemllm}, originally presented in Chinese and meticulously translated into English with careful attention to scientific accuracy and terminology. The physics questions are drawn from the \textbf{high-school-physics}~\cite{highschoolphysics2023} dataset, which includes a range of conceptual and quantitative exercises suitable for secondary-level learners. The mathematics component combines easy-level problems from the Illustrative Mathematics curriculum with medium- to hard-level questions selected from the \textbf{MATH-500}~\cite{lightman2023lets} dataset. 
Furthermore, each domain encompasses diverse sub-domains, collectively covering 15 comprehensive pedagogical scenarios, as illustrated in Figure~\ref{fig:distribution}.
All data sources were standardized into a unified format and consolidated to enable consistent and comprehensive evaluation across subjects.

\subsection{Evaluation Metric}
\label{subsec:metric}
In this subsection, we will detail the performance evaluation rubrics in \ours.

\noindent \textbf{Evaluation Dimensions.} 
To comprehensively evaluate the quality of generated visualizations in supporting student understanding and learning, we introduce a fine-grained scoring metric grounded in five pedagogically motivated dimensions:
\textbf{(1) Context Visualization}: evaluates how clearly the visualization situates the problem within a relevant context;
\textbf{(2) Diagram Design}: assesses the clarity, accuracy, and effectiveness of the diagrams used to represent information;
\textbf{(3) Text--Graphic Integration}: measures the coherence between textual explanations and visual elements, ensuring mutual interaction;
\textbf{(4) Thought Guidance}: examines the extent to which the visualization supports reasoning processes and highlights critical thinking steps;
\textbf{(5) Interactivity}: evaluates whether and how the visualization invites students engagement, reflection, or active manipulation.
Each dimension captures a distinct aspect of effective multimedia learning, with detailed rubrics provided in Appendix~\ref{app:visualdis} to guide the scoring process.

\begin{figure}[htbp]
    \centering
    \includegraphics[width=0.9\linewidth]{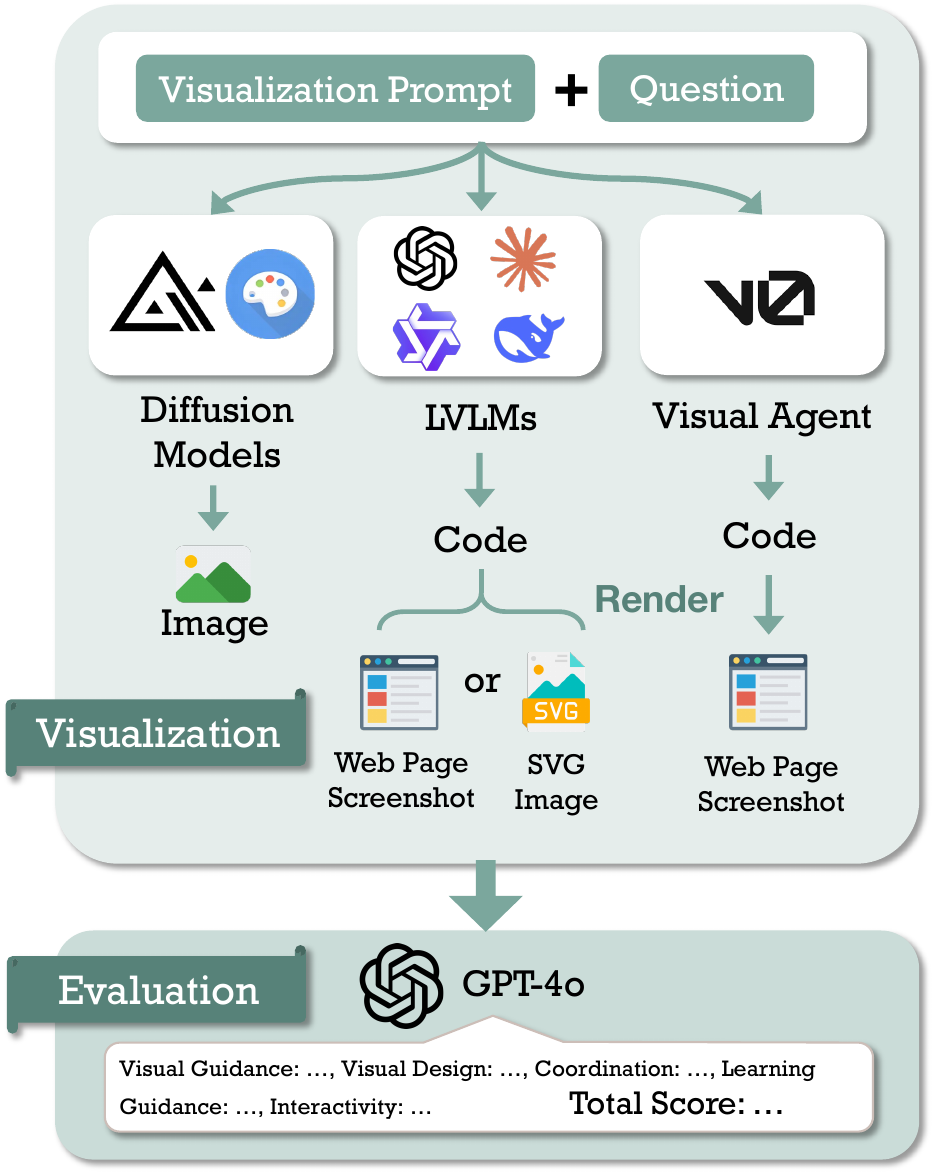}
    \caption{Workflow for the \ours\ benchmark evaluation. Models receive a visualization prompt and a question to generate visual outputs. All resulting visualizations undergo evaluation by GPT-4o across five dimensions to compute a total performance score.}
    \label{fig:workflow}
    \vspace{-1em}
\end{figure}

\noindent\textbf{Rubrics and Evaluation Procedure.}
To ensure a systematic and replicable evaluation pipeline, we manually construct detailed scoring rubrics for each of the five evaluation dimensions, with clear criteria defined for every level on a 0–5 scale. The visual outputs generated by each model are then assessed using GPT-4o within a controlled evaluation environment. For static image outputs from text-to-image models, the images are evaluated directly. For LVLMs, the generated SVG or HTML code is first rendered into its final visual form—either as a static image or an interactive webpage—before being evaluated. The detailed demonstration of our scoring procedure is in Figure \ref{fig:workflow}.

%% file: sec/multiagent.tex
\section{\ma}
\label{sec:multiagent}
Using the proposed \ours\ benchmark, we systematically evaluate the performance of existing text-to-image models and LVLMs (see detailed results in Table~\ref{tab:main_results} in Section~\ref{sec:experiments}). We find that most models perform poorly, with average scores below 50 on a 0–100 scale. This underperformance underscores the inherent challenge of decomposing complex reasoning and translating it into visual representations that align with human cognitive processes to effectively support education—a task that remains highly non-trivial.

To address these challenges, we propose a multi-agent system, \ma, inspired by pedagogical theories and designed to emulate the division of labor and collaborative reasoning found in expert instructional design. \ma\ consists of five specialized yet interdependent agents: a \textbf{Task Planning Agent}, which structures the instructional objective; a \textbf{Conceptual Mapping Agent}, which extracts and organizes key information; a \textbf{Reasoning Decomposition Agent}, which constructs step-by-step problem-solving logic; a \textbf{Metacognitive Reviewer}, which encourages summarization and learner reflection; and a \textbf{Visualization Agent}, which generates appropriate visual representations. This design introduces modularity and pedagogical interpretability by embedding distinct instructional roles directly into the agent workflow. The overall operation of \ma\ proceeds in two stages: (1) instructional flow construction and (2) collaborative solution generation, as detailed below.

\subsection{Instructional Flow Construction}
The first stage of \ma\ focuses on formulating a well-structured instructional task based on the original problem. A key challenge lies in analyzing the underlying reasoning structure, identifying implicit logical dependencies, and associating each reasoning step with relevant conceptual knowledge. To address this, we employ the \textbf{Task Planning Agent}, which systematically organizes the problem into an instructional format suitable for multimodal visualization. Its main functions include: (1) breaking down the problem into coherent subgoals, (2) clarifying the reasoning expected at each step, (3) aligning each step with domain-specific principles or formulas, and (4) anticipating potential student misconceptions or cognitive needs. This structured formulation provides a pedagogically grounded foundation that guides the downstream agents in generating coherent, targeted, and educationally effective visual explanations.

\subsection{Collaborative Solution Generation}

\begin{figure}
    \centering
    \includegraphics[width=0.8\linewidth]{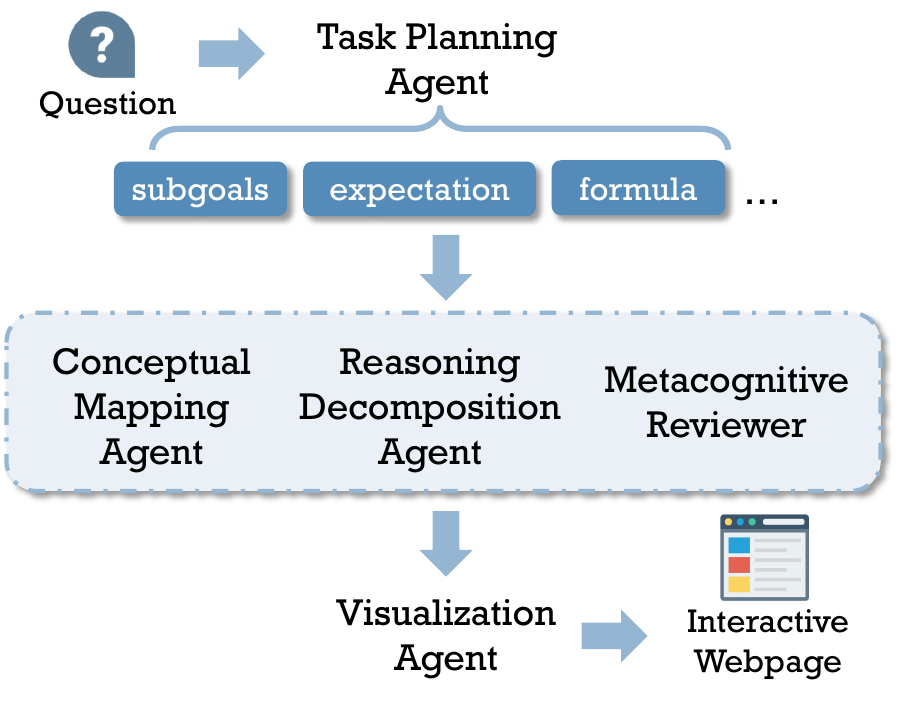}
    \caption{The structure of \ma.}
    \vspace{-1em}
    \label{fig:multiagent}
\end{figure}
In this stage, \ours\ executes the instructional task constructed by sequentially activating a set of specialized agents, each responsible for completing a specific aspect of the task. As shown in Figure \ref{fig:multiagent}, these agents operate in a coordinated manner to enhance the coherence of instructional logic, improving the clarity of visual representation, and ensure alignment with educational objectives. Specifically:

\noindent\textbf{Conceptual Mapping Agent.} This agent is responsible for extracting and organizing the core components of the input problem. Drawing on the Concrete–Representational–Abstract (CRA) instructional model~\cite{cramodel}, it classifies information into three categories: concrete entities, representational elements, and abstract constructs. The agent conducts fine-grained categorization and semantic summarization to support downstream visualization modules.

\noindent\textbf{Reasoning Decomposition Agent.} This agent decomposes complex problems into manageable subcomponents and provides step-specific instructional guidance. It applies the memory-oriented FOPS strategy~\cite{miller2020metacognitive}—find the problem type (e.g., equation solving, conceptual reasoning, commonsense application, or graphical interpretation), organize the structure via equations or diagrams, plan the solution path, and solve the task. Based on the decomposed steps, the agent also identifies critical instructional points that require additional support, especially those that benefit from visual scaffolding or interactive guidance.

\noindent\textbf{Metacognitive Reviewer.} Grounded in metacognitive theory~\cite{73ebe451-d5f5-3edf-b2bf-db7440f99ba5}, this agent supports learners in monitoring their comprehension and reasoning processes. It generates reflective prompts that foster self-questioning and self-correction, encouraging learners to evaluate the soundness of their problem-solving approaches.

\noindent\textbf{Visualization Agent.} This agent is responsible for constructing the “visual guidance” component of the instructional output. Instead of relying on decorative visuals, it emphasizes the use of abstract yet pedagogically effective representations—such as number lines, bar charts, schematic object illustrations, graphic organizers, sketch diagrams, and structured data tables. The agent ensures that each visualization is tightly aligned with the underlying abstract concept being taught. All visuals are rendered using the \texttt{v0}~\cite{v0_2025} system for web-based deployment.

%% file: sec/experiment.tex
\section{Experiments}
\label{sec:experiments}

\begin{table*}[htbp!]
\centering
\small
\caption{Performance of Diffusion Models, Large Vision Language Models and \texttt{v0} on \ours.}
\vspace{-0.3em}
\resizebox{\textwidth}{!}{
\begin{tabular}{l l | c c c | c c c | c c c | c}
\toprule
\multirow{2}{*}{\textbf{Method}} & \multirow{2}{*}{\textbf{Vis. Type}}
  & \multicolumn{3}{c|}{\textbf{Maths}}
  & \multicolumn{3}{c|}{\textbf{Physics}}
  & \multicolumn{3}{c|}{\textbf{Chemistry}}
  & \textbf{Avg} \\
       &                 
  & Easy & Medium & Hard
  & Easy & Medium & Hard
  & Easy & Medium & Hard
  &     \\
\midrule
\multicolumn{12}{c}{\textbf{Diffusion Model}} \\
\midrule
Flux.1-dev & Image & 13.8 & 13.4 & 13.2 & 11.7 & 8.5 & 10.0 & 20.0 & 16.6 & 16.0 & 13.8 \\
SD3.5 & Image & 17.3 & 20.3 & 18.8 & 16.8 & 13.0 & 12.0 & 22.8 & 21.7 & 34.0 & 18.4 \\
SDXL & Image & 17.3 & 23.3 & 25.5 & 18.9 & 15.4 & 24.0 & 33.6 & 30.2 & 24.0 & 21.8 \\
\midrule
\multicolumn{12}{c}{\textbf{Large Vision Language Model}} \\
\midrule
Deepseek VL2 & Webpage & 20.3 & 17.1 & 15.7 & 17.9 & 17.0 & 20.0 & 16.4 & 13.8 & 14.0 & 17.5 \\
GLM-4V-9B & Webpage & 22.3 & 21.1 & 19.4 & 24.5 & 21.5 & 24.0 & 22.3 & 21.5 & 16.0 & 21.9 \\
MiniCPM-V-2.6 & Webpage & 24.1 & 17.3 & 15.5 & 19.1 & 17.4 & 20.0 & 14.5 & 15.2 & 12.0 & 19.3 \\
Mistral-Small-3.1 & Webpage & 29.1 & 31.6 & 32.2 & 32.3 & 33.5 & 20.0 & 30.6 & 27.5 & 24.0 & 30.2 \\
Phi-3.5 & Webpage & 25.3 & 20.7 & 19.1 & 21.2 & 19.5 & 12.0 & 20.0 & 18.6 & 20.0 & 21.8 \\
Phi-4 & Webpage & 26.1 & 25.1 & 22.9 & 27.8 & 25.5 & 24.0 & 31.2 & 27.5 & 12.0 & 26.4 \\
Qwen2.5-VL-72B & Webpage & 24.3 & 18.1 & 15.8 & 19.7 & 17.1 & 24.0 & 18.2 & 16.4 & 12.0 & 20.0 \\
Claude 3.7 Sonnet & SVG & 61.2 & 26.7 & 23.6 & 18.5 & 16.9 & 14.0 & 47.5 & 47.2 & 18.0 & 42.0 \\
Claude 3.7 Sonnet & Webpage & 56.2 & 57.5 & 55.6 & 44.8 & 42.6 & 24.0 & 61.1 & 60.6 & 64.0 & 54.6 \\
GPT-4o & Webpage & 47.6 & 39.3 & 37.9 & 25.7 & 24.2 & 24.0 & 34.3 & 32.6 & 36.0 & 38.1 \\
GPT-4o & SVG & 36.1 & 19.7 & 19.5 & 13.0 & 12.8 & 4.0 & 30.0 & 27.5 & 22.0 & 26.3 \\
Gemini 2.0 Flash & Webpage & 46.9 & 9.5 & 15.7 & 31.7 & 26.5 & 24.0 & 32.0 & 25.8 & 30.0 & 43.6 \\
\midrule
\multicolumn{12}{c}{\textbf{Visualization Agent}} \\
\midrule
\texttt{v0} & Webpage & 63.0 & 37.6 & 47.2 & 53.3 & 58.5 & 52.0 & 74.7 & 52.8 & 68.0 & 58.2 \\
\bottomrule
\end{tabular}
}
\vspace{-1em}
\label{tab:main_results}
\end{table*}

This section outlines the experimental setup for benchmarking various foundation models on \ours. We evaluate Diffusion Models, LVLMs, a specialized visualization agent (\texttt{v0}), and our proposed \ma. Our investigation seeks to address the following key questions:
(1) How proficient are existing models at generating high-quality, explanatory visualizations within \ours?
(2) Can the proposed \ma\ system outperform current models?
(3) What distinct performance patterns emerge across different model architectures, academic disciplines, and evaluation dimensions in \ours?

\subsection{Experiment Setup}

\textbf{Baseline Models.} Our experimental evaluation encompasses a range of FMs, categorized as follows: (1) Image Generation Models: This category includes Flux.1-dev~\cite{flux2024}, Stable Diffusion 3.5 Large (SD3.5)~\cite{stability_introducing_stable_diffusion_3_5}, and Stable Diffusion XL Base 1.0 (SDXL)~\cite{Podell2023SDXLIL}. These models are tasked with generating static images directly from textual or visual inputs. (2) Large Vision-Language Models (LVLMs): We evaluate Deepseek-VL2~\cite{wu2024deepseekvl2mixtureofexpertsvisionlanguagemodels}, GLM-4V-9B~\cite{glm2024chatglm}, MiniCPM-V2.6~\cite{yao2024minicpm}, Mistral-Small-3.1-24B-Instruct-2503~\cite{mistral_small_3_1}, Phi-3.5-Vision-Instruct~\cite{Abdin2024Phi3TR}, Phi-4-Multimodal-Instruct~\cite{Abouelenin2025Phi4MiniTR}, Qwen2.5-VL-72B~\cite{qwen2.5-VL}, GPT-4o~\cite{hurst2024gpt}, Claude 3.7 Sonnet~\cite{anthropic_claude_3_7_sonnet}, and Gemini 2.0 Flash~\cite{google_gemini2_2025}. These models are prompted to generate SVG or HTML code, which is then rendered into visual outputs for evaluation. (3) Specialized Visualization Agent: We also assess \texttt{v0}~\cite{v0_2025}, an AI agent specifically designed to create interactive web pages based on instructional content.

\noindent\textbf{Evaluation Setups.} During evaluation, all generated visualizations are standardized into image format. For interactive web pages containing buttons, an automated script navigates through all accessible sub-pages, capturing individual screenshots of each. Performance is assessed using the evaluation metric described in Section~\ref{subsec:metric}, where GPT-4o scores the visual outputs based on predefined rubrics, assigning a score from 0 to 5 for each of the five dimensions. The cumulative score (maximum 25 points) is then normalized to a 0–100 scale for standardized reporting and comparison.

\subsection{Baseline Benchmarking}\label{sec:mainresults}

The performance of all evaluated baseline models is detailed in Table \ref{tab:main_results}. Across all evaluated models, the average scores indicate significant room for improvement. Diffusion Models generally exhibited the lowest performance, with average scores ranging from 13.8\% (Flux.1-dev) to 21.8\% (SDXL). This suggests that direct static image generation, while capable of producing visual elements, struggles substantially with the nuanced requirements of explanative and guiding visualizations for complex logical problems in our benchmark. 

LVLMs typically scored between 17.5\% (Deepseek VL2) and 30.2\% (Mistral-Small-3.1). Notable exceptions include Gemini 2.0 Flash (43.6\%) and Claude 3.7 Sonnet; the latter's significantly better performance with Webpages (54.6\%) over SVG (42.0\%). GPT-4o also showed a preference for Webpage generation (38.1\%) over SVG (26.3\%), suggesting that prompting advanced LVLMs for structured interactive webpages can yield more effective visual explanations. Nevertheless, even these top-tier LVLMs face considerable challenges in consistently meeting all of evaluation criteria. The visualization agent \texttt{v0}, specifically designed for webpage generation, achieved the highest average score among all baseline models at 58.2\%. This result highlights the advantage of a specialized agent in this task over more general-purpose FMs.

\begin{table*}[htbp!]
\centering
\resizebox{0.9\textwidth}{!}{
\begin{subtable}[ht]{0.47\linewidth}
\centering
\renewcommand{\arraystretch}{1.6}
\resizebox{0.95\textwidth}{!}{
\begin{tabular}{l l | c c c | c}
\toprule
\textbf{Method} & \textbf{Vis. Type} & \textbf{Easy} & \textbf{Medium} & \textbf{Hard} & \textbf{Avg} \\
\cmidrule(lr){1-2}\cmidrule(lr){3-5}\cmidrule(lr){6-6}
\multirow{6}{*}{\ma} & \multirow{6}{*}{Webpage} & \multicolumn{3}{c|}{\textbf{Maths}} & \multirow{6}{*}{\textbf{81.6}} \\
& & 90.2 & 64.5 & 65.0 & \\
\cmidrule(lr){3-5}
& & \multicolumn{3}{c|}{\textbf{Physics}} & \\
& & 85.3 & 81.7 & 84.0 & \\
\cmidrule(lr){3-5}
& & \multicolumn{3}{c|}{\textbf{Chemistry}} & \\
& & 69.0 & 76.3 & 76.0 &  \\
\bottomrule
\end{tabular}
}
    \caption{Performance of our \ma\ on \ours.}
    \label{tab:ma_results}
\end{subtable}
    \hfill
  %——右侧：Figure 4 ——
  \begin{subfigure}[ht]{0.52\linewidth}
  \centering
    \includegraphics[width=\linewidth]{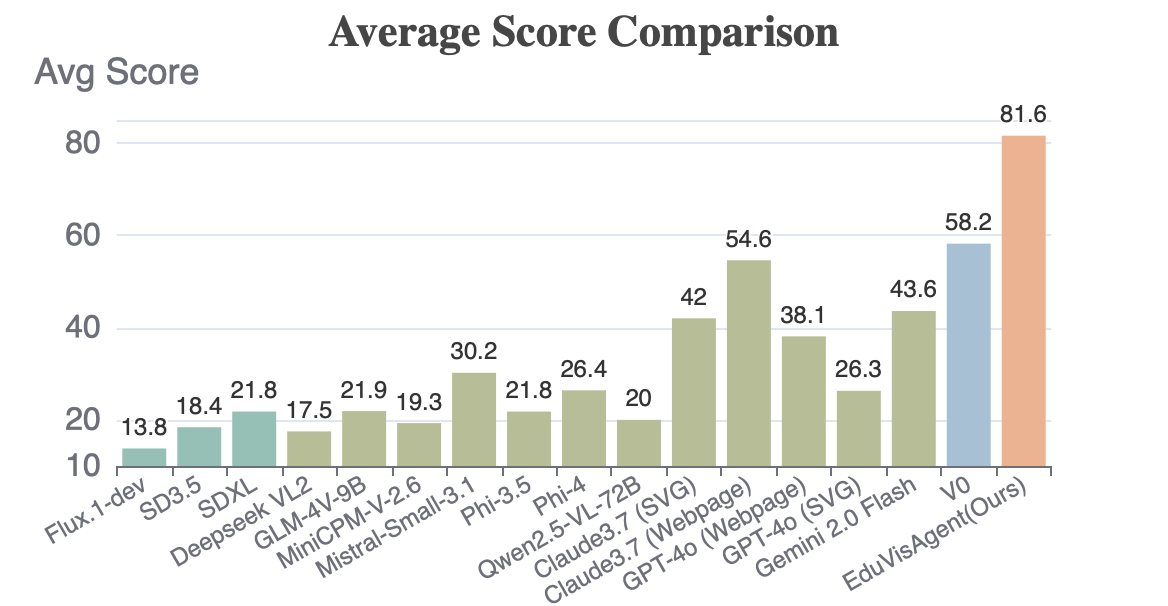}
    \caption{Comparison of average score across all models.
}
    \label{fig:barchart}
    \end{subfigure}
}
% \vspace{-0.7em}
\caption{Overall comparison of models: left is our \ma\ performance, right is the bar chart. \ma\ achieves the highest average score among all models.}
\label{tab:ma_performance}
% \vspace{-0.5em}
\end{table*}

\subsection{Performance Analysis of \ours}

Building upon the insights gained from the baseline evaluations, we assessed our proposed multi-agent system, \ma. The results, presented in Table \ref{tab:ma_performance} demonstrate a substantial leap in performance for generating explanative and logical valuable visualizations for STEM problems. \ma\ achieved an impressive overall average score of 81.6\%. Specifically, \ma\ surpasses the best-performing baseline \texttt{v0} (58.2\%), by a remarkable 23.4 percentage points. This constitutes an approximately 40.2\% relative improvement, underscoring the efficacy of our multi-agent architecture and the integration of educational methodologies. Compared to the best performing LVLM (Claude 3.7 Sonnet Webpage at 54.6\%) and the top diffusion model (SDXL at 21.8\%), the advancement offered by \ma\ is even more pronounced. These results clearly indicate that the design principles underlying \ma, which incorporate a multi-agent structure and pedagogical strategies, effectively address many of the limitations observed in existing generative models.

\begin{figure*}[h!]
    \centering
    \includegraphics[width=0.95\textwidth]{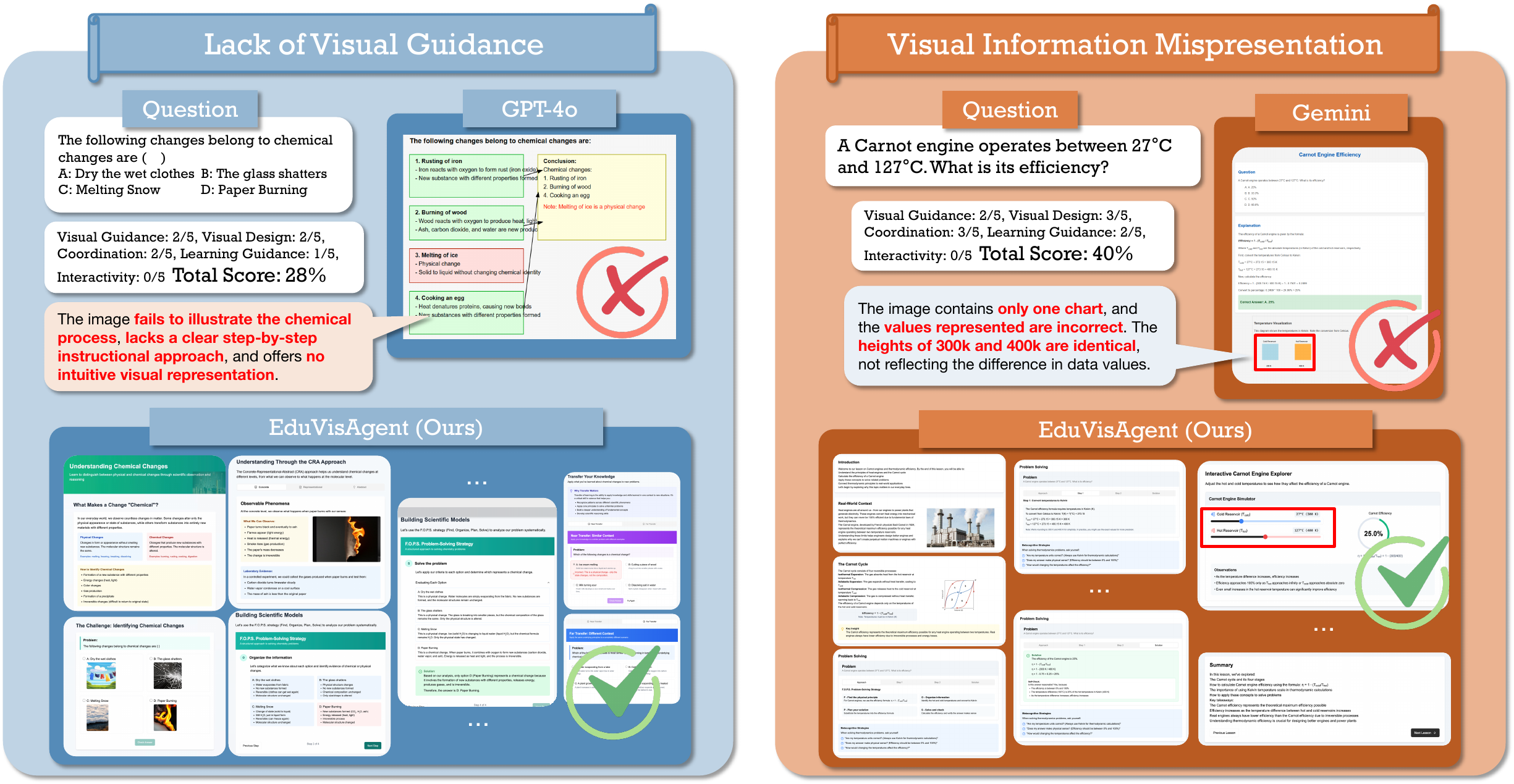}
    \caption{Baseline models versus our \ma. These examples clearly demonstrate the often poor output quality of baseline models, contrasting sharply with the high-quality, effective visualizations produced by \ma.}
    \label{fig:case_study}
    \vspace{-1em}
\end{figure*}

\subsection{Case Analysis}

To further illustrate the limitations of existing baselines and how our approach addresses these challenges, we present two case studies in Figure~\ref{fig:case_study}. On the left, for a chemistry question, the GPT-4o-generated solution lacks intuitive visualization of the chemical processes, resulting in fragmented information without visual guidance—reflected in a low score of just 28\%. In contrast, \ma\ begins by displaying background images of the relevant chemical elements, activating students’ prior knowledge. It then contextualizes each of the four answer options with real-world scenarios, thereby enhancing students' understanding of the underlying chemical transformations.

Conversely, for the Carnot cycle efficiency physics problem (right side of Figure \ref{fig:case_study}), the Gemini solution presents a single, flawed chart. Its depiction of 300K and 400K temperatures with identical heights introduces visual misinformation, failing to accurately represent data differences and thereby diminishing its pedagogical value. In stark contrast, \ma\ employs a multi-agent collaborative approach: it first generates a concrete factory scene to activate students' working memory of the "heat engine" concept. Subsequently, it constructs an accurate Carnot cycle diagram and offers a step-by-step problem breakdown, fostering clear conceptual understanding. Crucially, \ma\ provides interactive visualization components, enabling users to dynamically adjust temperatures via sliders and observe real-time changes in heat engine efficiency. This interactive engagement significantly facilitates higher-order thinking skills. 

Overall, through coordinated multi-agent optimization of image design, instructional structure, and learning pathways, \ma\ significantly outperforms traditional single-model approaches in accuracy, guidance, and interactivity.

\subsection{Fine-Grained Analysis on Five Evaluation Dimensions}
\begin{figure*}
    \centering
    \includegraphics[width=1\linewidth]{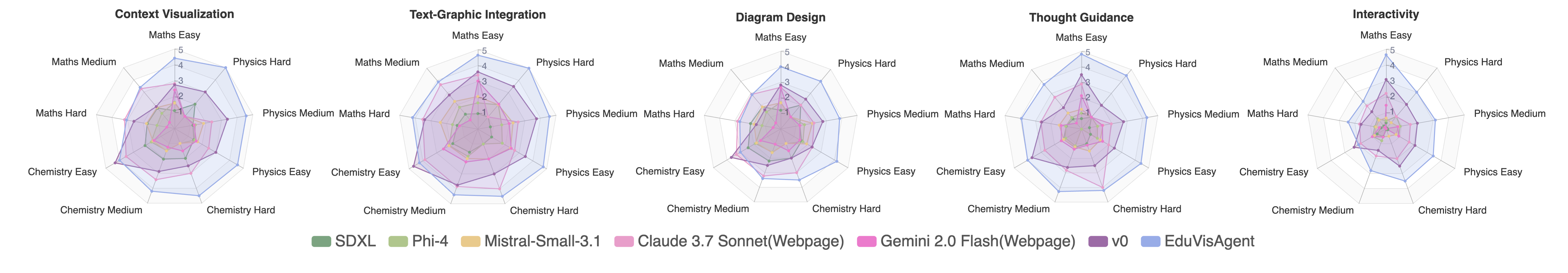}
    \vspace{-2em}
    \caption{Fine-grained performance comparison across our five key evaluation dimensions.}
    
    \label{fig:radar}
    \vspace{-1em}
\end{figure*}

Figure \ref{fig:radar} reveals distinct performance profiles for eight well performed evaluated models. In Context Visualization and Diagram Design, most baselines, including SDXL, Claude 3.7, and \texttt{v0}, exhibit moderate to low scores, often struggling with providing rich situational cues or pedagogically sound visual structures, especially for complex problems. \texttt{v0} and Claude show relatively better capabilities in Text-Graphic Integration and Thought Guidance compared to other FMs, which generally offer minimal support in these areas. However, all baseline models, including \texttt{v0}, are significantly limited in the Interactivity dimension, primarily due to their output format (static images/SVG or less dynamic webpages). In contrast, our \ma\ demonstrates consistently strong performance across all five dimensions. It particularly excels in creating rich context visualizations, well-structured diagram designs, and ensuring seamless text-graphic integration. Furthermore, \ma\ provides superior thought guidance and achieves notably high scores in Interactivity, areas where baseline models significantly lag. This comprehensive strength highlights \ma's advanced ability to generate not just visualizations, but truly effective and interactive pedagogical tools.

%% file: sec/related-work.tex
\vspace{-0.5em}
\section{Related Work}
\vspace{-0.5em}
\label{releted-work}

\noindent \textbf{LLM for Pedagogical Assistance.} Foundation models (FMs), including diffusion models and large vision-language models (LVLMs), have been increasingly adopted in educational contexts~\cite{chu2025llm, wang2024large} to support teaching and classroom interactions. EduAgent~\cite{xu2024eduagent} and Teachtune~\cite{jin2025teachtune} enhance the problem-solving process through automated simulations of student-teacher dialogues, collaborative learning, and task-oriented reasoning. Agents such as SEFL~\cite{zhang2025sefl} and PROF~\cite{nair2024closing} synthesize immediate, on-demand feedback to support large-scale instructional scenarios. Furthermore, domain-specific agents such as MathChat~\cite{wu2023mathchat}, NEWTON~\cite{wang2023newton}, and MEDCO~\cite{wei2024medco} further provide textual explanations tailored to scientific and medical education. While these systems address diverse pedagogical needs, their focus remains largely on text-based interactions~\cite{wu2023mathchat, xu2024eduagent, cui2024fine}, overlooking the critical role of visualization in fostering conceptual understanding and improving learning outcomes~\cite{presmeg2006research}. Despite its pedagogical importance, the capacity of FMs and agents to generate logical, explanatory visual illustrations remains underexplored. \ours\ is the first comprehensive benchmark designed to systematically evaluate FMs’ ability to produce pedagogically effective, step-by-step visual reasoning, covering 15 diverse visually grounded educational scenarios with multi-level problem sets and multimodal-centric solutions, providing a rigorous platform for visual pedagogy assessment.

\noindent \textbf{LLM for Scientific Visualization.} While some existing works have preliminarily explored the potential of FMs in supporting visual scaffolding~\cite{podo2024vi, chen2024halc,  pandey2025benchmarking, hong2025llms}, they are typically fragmented, lack pedagogical grounding, and fail to generalize across diverse educational tasks~\cite{wang2023llm4vis, ku2025theoremexplainagent}. For instance, Visual Sketchpad~\cite{hu2024visual} attempts to illustrate problem-solving processes with sketches generated from code. However, these visuals are often low in quality, lack logical coherence, and fall short in explanatory depth~\cite{wang2025multimodal}. Other approaches like MatplotAgent~\cite{yang2024matplotagent}, PlotGen~\cite{goswami2025plotgen}, and OmniSVG~\cite{yang2025omnisvg} leverage plotting and SVG tools to produce more accurate, data-grounded visualizations. Still, these methods are limited in scope, often addressing only isolated steps rather than providing systematic, end-to-end visual explanations of multi-step problem-solving tasks~\cite{vazquez2024llms, chen2024mj, chen2025shieldagent}. To overcome these limitations, we propose a multi-agent collaborative framework, \ma, that simulates the full learning journey—from initial problem exposure to deep conceptual understanding—by coordinating specialized agents to generate coherent, pedagogically aligned visualizations throughout the reasoning process.

%% file: sec/conclusion.tex
\vspace{-0.5em}
\section{Conclusion}
\vspace{-0.5em}
\label{conclusion}
This paper addressed the challenge of generating pedagogically meaningful visual explanations with AI systems. We introduced \ours, a benchmark revealing that existing models often produce inadequate visual outputs. To overcome this, we proposed \ma, a collaborative multi-agent framework. Experiments show \ma\ significantly outperforms all baselines, demonstrating the potential of agent-based systems for advancing educational visualization.

%% file: sec/appendix.tex
%%%%%%%%%%%%%%%%%%%%%%%%%%%%%%%%%%%%%%%%%%%%%%%%%%%%%%%%%%%%

\appendix

\newpage
\onecolumn
\section{Appendix}

\subsection{Visualization disciplines}\label{app:visualdis}

Table \ref{tab:discipline_visuals} illustrates the disciplines and types in our \ours.

\begin{table}[htb]
\centering
\begin{tcolorbox}[
    enhanced,
    boxrule=0pt,
    colback=gray!5,
    arc=3pt,
    shadow={1mm}{-1mm}{0mm}{black!10},
    width=0.98\columnwidth,
    top=3mm,
    bottom=3mm
]
\renewcommand{\arraystretch}{1.5}
\small
\centering
\begin{tabular}{
    >{\centering\arraybackslash}m{0.35\columnwidth}
    |
    m{0.54\columnwidth}
}
\hline
\textbf{Discipline} & \textbf{Common Visualization Types} \\ 
\hline\hline

\cellcolor{gray!10}
\textit{Mathematics} & 
Number lines, function graphs, and other formalized visual tools. \\ 
\hline

\cellcolor{gray!10}
\textit{Physics} & 
Diagrams involving levers, rigid body motion, forces, and fields. \\ 
\hline

\cellcolor{gray!10}
\textit{Chemistry} & 
Molecular structures and schematic representations of standard laboratory apparatus. \\ 
\hline

\end{tabular}
\end{tcolorbox}
\caption{Representative Visualization Types Across Academic Disciplines}
\label{tab:discipline_visuals}
\end{table}

\subsection{Evaluation Metric}
\paragraph{Visual Scenario Design Guidance}  
The category of "Visual Scenario Design Guidance" outlines different levels of visualizing mathematical concepts, progressing from basic text-only representations to highly integrated visual-text formats. Through five defined levels, the framework demonstrates how visual elements can enhance students' understanding and engagement with abstract ideas, guiding instructional designers to gradually enrich scenarios, add annotations, and strengthen contextual connections—ultimately achieving the goal of visually presenting the full flow and conceptual structure of the content.The five levels of Visual Scenario Design Guidance are as follows:

\begin{table}[htb]
\centering
\begin{tcolorbox}[
    enhanced,
    boxrule=0pt,
    colback=gray!5,
    arc=3pt,
    shadow={1mm}{-1mm}{0mm}{black!10},
    width=0.98\columnwidth,
    top=3mm,
    bottom=3mm
]
\renewcommand{\arraystretch}{1.5}
\small
\centering
\begin{tabular}{
    >{\centering\arraybackslash}m{0.15\columnwidth}
    |
    m{0.74\columnwidth}
}
\hline
\textbf{Level} & \textbf{Description} \\
\hline\hline

\cellcolor{gray!10}
Level 1 &
The image contains no scenes or illustrations, presenting only text and formulas. It lacks contextual visual cues, failing to spark interest or connect the concepts to real-life situations. \\
\hline

\cellcolor{gray!10}
Level 2 &
The image includes a single static illustration or low-fidelity mockup with minimal labeling that does not highlight variables or key objects, offering limited context and poor immersion. \\
\hline

\cellcolor{gray!10}
Level 3 &
Multiple static schematic diagrams or sketch-style illustrations appear in the image, labeling core objects, variables, and simple steps, providing basic visual guidance but lacking layered coherence. \\
\hline

\cellcolor{gray!10}
Level 4 &
The image integrates scenario illustrations, storyboard panels, and infographics to present the process in multiple views and steps, with annotations and captions guiding students through mapping abstract concepts to context. \\
\hline

\cellcolor{gray!10}
Level 5 &
Storyboard-style illustrations and infographics are fused into a single image, including overview, detailed close-ups, and key pathway diagrams with comprehensive annotations, allowing students to grasp the entire flow and conceptual network at a glance. \\
\hline

\end{tabular}
\end{tcolorbox}
\caption{Five Levels of Visual Scenario Design Guidance}
\label{tab:visual_scenario_guidance}
\end{table}

\paragraph{Visual Illustration Design}  
The category of "Visual Illustration Design" describes progressive levels of visual elements used to support students’ systematic understanding of quantities and relationships. It ranges from no visual aids to complex integrated dashboards that deeply connect data and model structures. Through five levels, the framework guides designers to improve clarity, coherence, and contextual richness of visual illustrations, enhancing students’ analytic and comparative abilities.

\begin{table}[htb]
\centering
\begin{tcolorbox}[
    enhanced,
    boxrule=0pt,
    colback=gray!5,
    arc=3pt,
    shadow={1mm}{-1mm}{0mm}{black!10},
    width=0.98\columnwidth,
    top=3mm,
    bottom=3mm
]
\renewcommand{\arraystretch}{1.5}
\small
\centering
\begin{tabular}{
    >{\centering\arraybackslash}m{0.15\columnwidth}
    |
    m{0.74\columnwidth}
}
\hline
\textbf{Level} & \textbf{Description} \\
\hline\hline

\cellcolor{gray!10}
Level 1 &
The image contains no charts, axes, or flow diagrams—only text. Without embedded visual tools, students cannot systematically organize or analyze quantities and relationships. \\
\hline

\cellcolor{gray!10}
Level 2 &
The image presents a static number line and colored bar chart with complete scales and a legend, helping students gain a basic understanding of numerical changes. However, it lacks comparison and contextual layering.
\\
\hline

\cellcolor{gray!10}
Level 3 &
The image presents a static number line and colored bar chart with complete scales and legends, helping students grasp basic numerical changes visually, though comparison and context layering are absent. \\
\hline

\cellcolor{gray!10}
Level 4 &
The image combines number lines, flowcharts, infographics, and arrow annotations; multiple visuals are juxtaposed or overlaid to show processes and variable changes for a coherent modeling view. \\
\hline

\cellcolor{gray!10}
Level 5 &
The image presents a dashboard-style visualization integrating axes, bar charts, flow diagrams, heatmaps, etc., with linked elements that deeply visualize data relationships and model structure. \\
\hline

\end{tabular}
\end{tcolorbox}
\caption{Five Levels of Visual Illustration Design}
\label{tab:visual_illustration_design}
\end{table}
\paragraph{Text–Illustration Coordination}  
The category of "Text–Illustration Coordination" describes levels of alignment and integration between textual content and visual elements within images. This progression ranges from complete disconnection to seamless fusion, enabling students to effectively map and synthesize text, formulas, and graphics. The framework guides designers in strengthening links between verbal and visual information to enhance comprehension and structural understanding.

\begin{table}[htb]
\centering
\begin{tcolorbox}[
    enhanced,
    boxrule=0pt,
    colback=gray!5,
    arc=3pt,
    shadow={1mm}{-1mm}{0mm}{black!10},
    width=0.98\columnwidth,
    top=3mm,
    bottom=3mm
]
\renewcommand{\arraystretch}{1.5}
\small
\centering
\begin{tabular}{
    >{\centering\arraybackslash}m{0.15\columnwidth}
    |
    m{0.74\columnwidth}
}
\hline
\textbf{Level} & \textbf{Description} \\
\hline\hline

\cellcolor{gray!10}
Level 1 &
Text and illustrations in the image are completely disconnected, with no labels, legends, or connectors—students cannot use visuals to understand text or formulas. \\
\hline

\cellcolor{gray!10}
Level 2 &
Text occasionally prompts “see diagram” or “refer to the illustration,” but the image lacks legends or clear labels, so mapping between text and graphics remains ambiguous. \\
\hline

\cellcolor{gray!10}
Level 3 &
Text descriptions and image elements share consistent numbering, color blocks, or arrows linked to a simple legend, explaining core symbols and variables to support initial mapping. \\
\hline

\cellcolor{gray!10}
Level 4 &
Text paragraphs are laid out alongside corresponding visuals within the same image, with detailed legends and color-coded annotations enabling simultaneous reading and mapping. \\
\hline

\cellcolor{gray!10}
Level 5 &
Text, formulas, and legends are fully integrated in one image, using consistent colors, numbering, and layered layout to achieve seamless text–graphic fusion for complete structural understanding. \\
\hline

\end{tabular}
\end{tcolorbox}
\caption{Five Levels of Text–Illustration Coordination}
\label{tab:text_illustration_coordination}
\end{table}

\paragraph{Learning Thought Guidance}  
The category of "Learning Thought Guidance" describes the progressive inclusion of visualized problem-solving strategies and reflective cues in images. From presenting only problem statements to complex integrated dashboards, this framework guides designers to scaffold students’ strategic thinking and metacognitive reflection through visual tools, enabling deeper reasoning and transfer of learning.

\begin{table}[htb]
\centering
\begin{tcolorbox}[
    enhanced,
    boxrule=0pt,
    colback=gray!5,
    arc=3pt,
    shadow={1mm}{-1mm}{0mm}{black!10},
    width=0.98\columnwidth,
    top=3mm,
    bottom=3mm
]
\renewcommand{\arraystretch}{1.5}
\small
\centering
\begin{tabular}{
    >{\centering\arraybackslash}m{0.15\columnwidth}
    |
    m{0.74\columnwidth}
}
\hline
\textbf{Level} & \textbf{Description} \\
\hline\hline

\cellcolor{gray!10}
Level 1 &
The image offers no visualized problem-solving guidance, showing only the problem statement and formulas, leaving students without strategic cues or reflection prompts. \\
\hline

\cellcolor{gray!10}
Level 2 &
The image embeds a simple flowchart or two title-style hints (e.g., “Identify problem type,” “Check result”), but the flowchart is overly simplistic and hints lack hierarchical detail. \\
\hline

\cellcolor{gray!10}
Level 3 &
The image displays a step-by-step flowchart template with key thinking nodes and self-check checkpoints, leaving annotation space for students to visually record their reasoning. \\
\hline

\cellcolor{gray!10}
Level 4 &
The image combines a near-transfer exercise with a comparative thought diagram, visually highlighting strategy differences so students can apply existing reasoning to a new context. \\
\hline

\cellcolor{gray!10}
Level 5 &
The image fuses near- and far-transfer exercises, concept mind maps, and a reflection panel into a dashboard-style layout, allowing students to review and extend their problem-solving network visually. \\
\hline

\end{tabular}
\end{tcolorbox}
\caption{Five Levels of Learning Thought Guidance}
\label{tab:learning_thought_guidance}
\end{table}

\paragraph{Interactivity and Personalized Support}  
The category of "Interactivity and Personalized Support" outlines levels of incorporating feedback, hints, and tailored assistance into images, evolving from static presentations to dynamic, student-responsive visual supports. This framework encourages designers to embed interactive elements that adapt to learner needs, promoting engagement and personalized problem-solving.

\begin{table}[htb]
\centering
\begin{tcolorbox}[
    enhanced,
    boxrule=0pt,
    colback=gray!5,
    arc=3pt,
    shadow={1mm}{-1mm}{0mm}{black!10},
    width=0.98\columnwidth,
    top=3mm,
    bottom=3mm
]
\renewcommand{\arraystretch}{1.5}
\small
\centering
\begin{tabular}{
    >{\centering\arraybackslash}m{0.15\columnwidth}
    |
    m{0.74\columnwidth}
}
\hline
\textbf{Level} & \textbf{Description} \\
\hline\hline

\cellcolor{gray!10}
Level 1 &
The image includes no feedback or support components—only a static problem statement and answer field—offering no hints, examples, or error cues and resulting in a nonresponsive visual. \\
\hline

\cellcolor{gray!10}
Level 2 &
The image shows fixed hint boxes (e.g., “Hint: draw a number line,” “Hint: check rounding”), but hints are not tailored to student responses, limiting personalized guidance. \\
\hline

\cellcolor{gray!10}
Level 3 &
The image integrates multiple static correction tips and example solution modules (common mistakes and standard approaches), which students can reference visually but without intelligent recommendations. \\
\hline

\cellcolor{gray!10}
Level 4 &
The image presents example solution workflows, text hints, and a common-errors analysis section highlighted with color blocks and arrows, providing diverse visual support in a single layout. \\
\hline

\cellcolor{gray!10}
Level 5 &
The image displays a comprehensive visual support panel with difficulty suggestions, personalized hints, worked examples, and extension resource links, enabling students to select tailored guidance directly from the visual layout. \\
\hline

\end{tabular}
\end{tcolorbox}
\caption{Five Levels of Interactivity and Personalized Support}
\label{tab:interactivity_personalized_support}
\end{table}

\subsection{Similarity between GPT and Human Evaluation}

\begin{table*}[htbp!]
\small
\centering
\renewcommand{\arraystretch}{1.6}
\resizebox{0.6\textwidth}{!}{
\begin{tabular}{l | c c c | c}
\toprule
\textbf{Metric} & \textbf{Chemistry} & \textbf{Math} & \textbf{Physics} & \textbf{Average} \\
\midrule
Cosine Similarity ↑ & 0.9742 & 0.9557 & 0.9666 & \textbf{0.9655} \\
MSE ↓               & 0.3895 & 0.7093 & 0.6118 & \textbf{0.5702} \\
\bottomrule
\end{tabular}
}
\caption{Cosine similarity and mean squared error across subjects. Math is the average of Math500 and IllustrativeMath, each with 50 samples.}
\label{tab:cosine_mse_summary}
\end{table*}

We evaluated the similarity between GPT-based evaluations and human evaluations to assess the reliability of GPT's scoring capabilities. Specifically, we selected 50 samples from each subject category (Chemistry, Math, and Physics) and had both GPT and human evaluators independently rate these samples. Our human evaluators were undergraduate students from top universities, ensuring high-quality and informed assessments. We measured agreement using two standard metrics: Cosine Similarity and Mean Squared Error (MSE). As shown in Table~\ref{tab:cosine_mse_summary}, the high Cosine Similarity scores (averaging 0.9655) and low MSE values (averaging 0.5702) across all subjects indicate that GPT’s scoring aligns closely with human judgment. These results demonstrate that GPT-based evaluation is highly reliable and sufficiently robust for practical usage, closely approximating human evaluative standards.

\subsection{Evaluation Prompt}
The instructional web page evaluation prompt is structured as follows:

\begin{center}
\begin{tcolorbox}[enhanced, colback=white!5!, colframe=gray!80!violet!90!, boxrule=0.5mm, arc=3pt, width=0.98\columnwidth, title=Evaluation Prompt]

As a \textbf{professional evaluator of instructional web pages}, your task is to determine whether the generated web page meets expectations across five specific categories.

\vspace{0.5em}
\textbf{Instructions:}
\begin{itemize}
  \item Assign an \textbf{integer score from 0 to 5} for each of the five categories (1–5).
  
  \item \textbf{0} = \textcolor{red!70!black}{completely missing or extremely poor} \\
        \textbf{5} = \textcolor{green!50!black}{fully meets the highest standard}
  
  \item Evaluation should be based solely on the specified aspect: \textit{\{category\}}. \\
        The definition of \textit{\{category\}} is: \textit{\{description\}}.
  
  \item \textcolor{red!80!black}{\textbf{Do not include any explanation, justification, or additional commentary. Refusing to provide a score is not allowed.}}
\end{itemize}

\end{tcolorbox}

\vspace{1em}

\begin{tcolorbox}[colback=white!10, colframe=gray!85!blue!90, boxrule=0.5mm, arc=3mm, width=0.98\columnwidth, title=Evaluation Output Format]
\texttt{\{\{RATING: \{"1":score, "2":score, "3":score, "4":score, "5":score\}\}\}}
\end{tcolorbox}
\end{center}

\subsection{Additional Related Works}

\textbf{LLM-based AI Agents.} Recent advancements in LLM-based agents have led to the development of specialized architectures capable of long-horizon planning, tool use, and memory management across a range of real-world domains~\cite{yao2023react, chan2024mle, chen2024agentpoison,chen2025celltypeagent,nie2025weak,han2025mdocagent,zhou2025anyprefer}. In software engineering, agents like Devin~\cite{cognition2024devin}, CodeAgent~\cite{zhang2024codeagent}, and SWE-agent~\cite{jimenez2023swe} manage full development pipelines and perform iterative code debugging. In the domain of web automation, agents have been deployed for complex web navigation and interaction tasks~\cite{wang2024agent, chen2025shieldagent, zhou2023webarena}. LLM-based agents have also demonstrated effectiveness in embodied settings such as robotic manipulation, autonomous driving, and embodied navigation~\cite{song2023llm, shridhar2020alfworld, mao2023language,yuan2025remac}. Beyond these, specialized agents have emerged in domains including healthcare~\cite{qiu2024llm}, finance~\cite{yu2024finmem}, and academic research~\cite{starace2025paperbench}. 

In the educational domain, AI agents such as EduAgent~\cite{xu2024eduagent} and Teachtune~\cite{jin2025teachtune} simulate student-teacher dialogues, collaborative learning activities, and task-oriented reasoning to enhance problem-solving instruction. Agents like SEFL~\cite{zhang2025sefl} and PROF~\cite{nair2024closing} generate on-demand feedback for large-scale educational settings, while domain-specific tools such as MathChat~\cite{wu2023mathchat}, NEWTON~\cite{wang2023newton}, and MEDCO~\cite{wei2024medco} provide textual explanations for scientific and medical learning.

Despite these advancements, few works explore collaborative, multi-agent designs tailored for educational reasoning and visualization. \ma\ is the first systematic multi-agent framework that coordinates specialized agents for instructional planning, reasoning decomposition, metacognitive prompting, and visualization design, offering a comprehensive approach to support step-by-step pedagogical problem-solving.

%%%%%%%%%%%%%%%%%%%%%%%%%%%%%%%%%%%%%%%%%%%%%%%%%%%%%%%%%%%%